\RequirePackage{fix-cm}
\documentclass[twocolumn]{svjour3}          
\smartqed  
\usepackage{graphicx}
\usepackage{mathptmx}      
\usepackage{amsmath}
\usepackage{multirow}
\usepackage{float}
\newcommand{\eprod}[2]{\left(#1,#2 \right)}
\newcommand{\neprod}[3][]{\eprod{#2}{#3}_{#1}}
\newcommand{\pc}[1]{\varphi_{#1}}
\newcommand{\pcset}[1]{\{ \pc{i} \}_{i=1}^{#1}}
\newcommand{\pcsethat}[1]{\{ \hat{\pc{i}} \}_{i=1}^{#1}}
\newcommand{\trunc}[2]{#1_{#2}}
\newcommand{\resid}[2]{#1_{#2 r}}
\newcommand{\norm}[2][]{||#2||_{#1}}
\newcommand{\mean}[1]{<#1>}
\newcommand{\nexp}[2]{#1\times10^{#2}}

\journalname{ArXiv}
\begin{document}

\title{Objective-Sensitive 
       Principal Component Analysis \\
       for High-Dimensional Inverse Problems
}


\author{Maksim Elizarev   \and
        Andrei Mukhin     \and
        Aleksey Khlyupin
}


\institute{ M. A. Elizarev \and A. V. Mukhin \and A. N. Khlyupin \at
Moscow Institute of Physics and Technology
\and
M. Elizarev (corresponding author) \at
\email{elizarev@phystech.edu}
\and
A. Mukhin \at
\email{andrei.mukhin@phystech.edu} 
\and
A. Khlyupin \at
\email{khlyupin@phystech.edu} 
}


\maketitle

\begin{abstract}
We present a novel approach for adaptive, differentiable parameterization of large-scale random fields. If the approach is coupled with any gradient-based optimization algorithm, it can be applied to a variety of optimization problems, including history matching. The developed technique is based on principal component analysis (PCA) but modifies a purely data-driven basis of principal components considering objective function behavior. To define an efficient encoding, Gradient-Sensitive PCA uses an objective function gradient with respect to model parameters. We propose computationally efficient implementations of the technique, and two of them are based on stationary perturbation theory (SPT). Optimality, correctness, and low computational costs of the new encoding approach are tested, verified, and discussed. Three algorithms for optimal parameter decomposition are presented and applied to an objective of 2D synthetic history matching. The results demonstrate improvements in encoding quality regarding objective function minimization and distributional patterns of the desired field. Possible applications and extensions are proposed.
\end{abstract}

\section{Introduction} \label{intro}
Inverse problems appear in many areas of comparative research, where the problem of defining uncertain inner properties is considered having a set of its life-cycle observations. Although a correct solution for a lot of practical inverse problems requires an efficient parametrization algorithm, our specific interest lies in the field of history matching problem. The purpose of this procedure is to generate a detailed reservoir description consistent with prior information and match production data to within some tolerance. History matching is usually done using two types of data, namely static and dynamic. Static data is mostly constant over time, e.g., a geological concept of formation, well logs, and petrophysical data and is commonly given as prior information. Dynamic data is time-dependent and represents properties change during a production process, e.g., pressure and flow rates, flow responses. A relatively recent review on the history matching problem can be found in \cite{oliver2011recent}.

A common approach is to perform history matching in the optimization framework or as a data assimilation problem. For the latter, ensemble methods, such as ensemble Kalman filters (EnKF) \cite{aanonsen2009ensemble,evensen2007using}, Ensemble Smoother \cite{skjervheim2011ensemble,chen2014history} recently have gained popularity. Such methods require black-box access regarding a forward flow simulator and provide multiple results. These features simplify the process of uncertainty quantification and allow working with a black-box simulator \cite{hajizadeh2011ant,hajizadeh2011towards}. However, an ensemble collapse problem is a commonly occurring phenomenon that limits performance. This circumstance leads to a need for a large number of members within an ensemble and high computational costs.

In optimization context, the history matching problem is often addressed by stochastic methods such as genetic algorithm\cite{romero2000modified}, particle swarm optimization\cite{mohamed2010application,hajizadeh2011towards}, evolutionary algorithms\cite{hajizadeh2010history,schulze2001optimization} and others\cite{hajizadeh2011ant}. Although these methods perform a global search and allow using a forward simulator as a black box, their use for complex models has a relatively high computational cost and can be applied only via decent computational clusters.

Originally, this work is focused on gradient optimization methods. These methods are invasive w.r.t the forward simulator and provide a local search, but they are sufficiently faster than the methods described above \cite{kaleta2011model,sarma2006efficient,jansen2011adjoint}. Adjoint-based techniques are usually applied in history matching procedure in this context since they provide required gradients at a computational cost of one additional forward simulation. These techniques are investigated for partial differential equations (PDE), e.g. with applications in closed-loop reservoir management \cite{jansen2011adjoint,sarma2006efficient,van2012adjoint}, and even for integrodifferential equations (IDE) of systems with a memory effect\cite{eage:/content/papers/10.3997/2214-4609.201802214}.

Since history matching often has to be performed on the real large-scale fields, parametrization techniques are useful and needful as they substantially reduce the number of parameters that have to be determined. Also, it allows maintaining geological consistency of a result. Some deterministic approaches, such as discrete cosine transform (DCT) \cite{jafarpour2008history} or discrete wavelet transform (DWT) \cite{sahni2005multiresolution} allow representing a model in terms of relatively few parameters, but their performance on complex models is inappropriate. That is the case since information about a geological model's prior covariance is not considered in a subspace construction process. On the other hand, one may use PCA-based algorithms. Classic linear PCA, also known as Karhunen -- Loeve expansion or Proper Orthogonal Decomposition (POD), was successfully applied to history matching problem \cite{sarma2006efficient}. Since classic PCA considers only the covariance matrix of a data, it preserves only two-point statistics, and several approaches were developed for efficient encoding of complex non-Gaussian fields. Among them are kernel PCA (kPCA) \cite{sarma2008kernel}, optimization PCA (O-PCA) \cite{vo2014new}, regularized kPCA (R-kPCA) \cite{vo2016regularized}, convolutional neural network PCA (CNN-PCA) \cite{liu2018deep}.

\paragraph{Research suggestion.} 
A drawback of existing parameterization techniques is their pure data-driven nature. The quality metrics of the mentioned methods measure a loss caused by a projection of dataset points onto derived subspaces. While solving an inverse problem, this circumstance leads to a significant constraint of a search area without any confidence that the chosen subspace meets requirements for representativeness in terms of objective function minimization.
Some patterns, sufficiently affecting objective function value, can be truncated within principal components deemed sufficient in terms of a given dataset. For example, there is no guarantee that objective function loss caused by a projection decreases significantly or even monotonically with an increase of subspace dimensionality. Furthermore, if a given data is of high uncertainty and low quality, it is not consistent with the real properties of a studied object, and essential patterns can be underrepresented in the dataset. In this work, we look forward to overcoming such issues.

\begin{figure}[b]
\centering
\includegraphics[width = 1\linewidth]{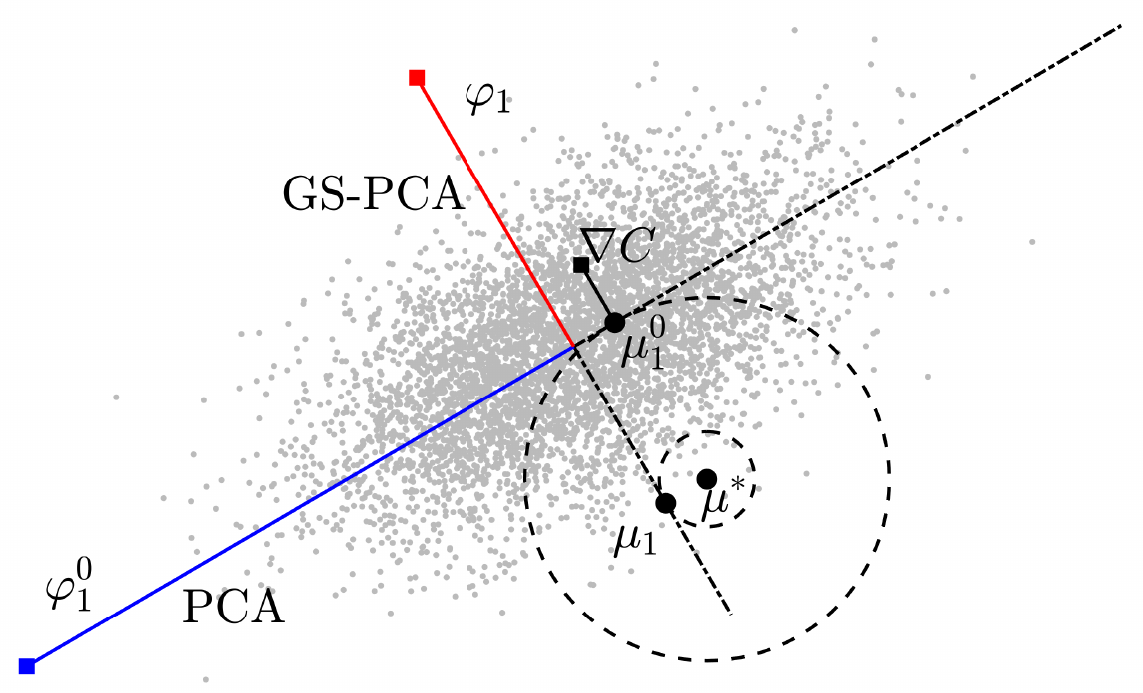}
\caption{Geometric explanation of Gradient-Sensitive PCA compared to classic PCA. The first principal component $\pc{1}$ accounts for both grey-colored data points and an objective with a global minimum $\mu^*$ and gradient $\nabla C$ in a trial point $\mu_1^0$, which is a constrained minimum in a subspace of the first classic principal component $\pc{1}^0$. A constrained minimum in a subspace of gradient-sensitive component is indicated as $\mu_1$. All the displayed vectors were properly generated by the corresponding algorithms for a parabolic objective function.}
\label{fig:toc}
\end{figure}

The general proposition of this paper is to include an objective function in the quality metrics of parameterization. Correspondingly, we present a set of novel approaches. Conceptual visualization of gradient-sensitive PCA (GS-PCA) is depicted in Fig.~\ref{fig:toc}. The key idea is to account for local objective gradient $\nabla C$  to derive a subspace of principal components $\varphi$ more descriptive than pure data-driven components in terms of an objective function error caused by parameterization of a field $\mu$.
In addition to the new algorithms of parameterization, we propose a computationally efficient algorithm for selecting more suitable principal components from among standard ones.

\paragraph{Contents.}
This paper is arranged as follows. First of all, a notation for PCA and history matching problem is introduced in Section~\ref{sec:methods}. Next, that section devoted to a formulation for the novel optimal gradient-sensitive decomposition, corresponding exact solution, and two approximate approaches based on stationary perturbation theory (SPT). In Section~\ref{sec:results}, some train and test scores of all three GS-PCA algorithms are given, as well as additional visual representations.  Consequently, Section~\ref{sec:disc} contains an analysis of proposed solutions' accuracy, observed performance, advances, and some issues. Eventually, we discuss some possible application approaches of the GS-PCA, highlighting research opportunities.

\section{Methods} \label{sec:methods}
In this paper, we operate with Euclidean scalar product $\neprod{x}{y}$ of arbitrary vectors $x$ and $y$ 
\begin{equation}
    \eprod{x}{y}=x^Ty=  \int_{r} x(r) y(r) dr
\end{equation}
and non-Euclidean scalar product $\neprod[\Theta]{x}{y}$ with corresponding operator $\Theta$. 
\begin{equation}
    \neprod[\Theta]{x}{y}=x^T \Theta y = \int_{r_1} \int_{r_2} x(r_1) \Theta(r_1,r_2) y(r_2) dr_2 dr_1
\end{equation}
We imply either discrete or continuous form for vectors and operators and the corresponding area of applicability.

\subsection{PCA} \label{sec:pca}
Let us consider a decomposition of a vector $\mu$ over an orthonormal basis $\{\pc{i}\}$.
\begin{gather}\label{eq:param-decomp}
    \mu = \sum_{i} a_i \pc{i} \\
    \neprod[\Theta]{\pc{i}}{\pc{j}}=\delta_{ij} \\
     a_i = \neprod[\Theta]{\pc{i}}{\mu}
\end{gather}
This decomposition can be split into a truncated decomposition $\trunc{\mu}{N}$ and a residual term $\resid{\mu}{N}$.

\begin{gather}\label{eq:trunc-resid}
    \trunc{\mu}{N}=\sum_{i=1}^N a_i \pc{i} \\
    \resid{\mu}{N} = \mu - \trunc{\mu}{N}=\sum_{i>N} a_i \pc{i}
\end{gather}
The principal components $\pcset{N}$ of a dataset minimize the mean Euclidean scalar square of the residual $\mean{\norm{\resid{\mu}{N}}^2}$ over any given dataset and number of orthonormal principal components $N$.
\begin{align}\label{eq:pca-problem}
    \begin{aligned}
         &\pcset{N} = \arg \min_{\pc{}} \mean{\norm{\resid{\mu}{N}}^2}~\forall N \\ 
         &\text{s.t.}~\eprod{\pc{i}}{\pc{j}} = \delta_{ij} 
    \end{aligned}
\end{align}

\begin{equation}\label{eq:enorm}
   \norm{x}^2 = \eprod{x}{x}
\end{equation}
The necessary condition of constrained extremum leads to the problem of eigenvectors and eigenvalues of the second-moment operator $K$ \cite{schmidt1989theorie}. To be concise, we refer to the second moment K as to a covariance.
\begin{gather} \label{eq:eigen}
K(x,y) = \mean{\mu(x) \mu(y)},~K^T = K\\
    K\pc{k}=\sigma_k \pc{k}
\end{gather}
The singular value decomposition (SVD) provides a solution to the eigenproblem in a finite case.
\begin{align}\label{eq:pca-svd}
    \begin{aligned}
        & K = \Phi \Sigma^2 \Phi^T \\
        & \Phi = [\pc{1},\pc{2},\dots] \\
        & \Sigma^2 = \text{diag}(\sigma_1,\sigma_2,\dots) ,~\sigma_1 \geq \sigma_2 \geq \dots 0
    \end{aligned}
\end{align}

\subsection{Objective-Sensitive PCA (OS-PCA)} \label{sec:os-pca}
A history matching problem implies calculation of a parameter vector $\mu^*$ which provides a hydrocarbon recovery simulation $S(\mu)$ close to prior observations $S_0$. $C(\mu)$ is a corresponding objective function to be minimized.
\begin{gather}
    \label{eq:matching}
    \mu^*=\arg \min_{\mu} C(\mu) \\
    C(\mu) = \norm{S(\mu) - S_0}^2
\end{gather}
The general proposition of this paper is to consider an original condition of decomposition optimality for some non-Euclidean set of orthonormal principal components regularized with objective function residual $\resid{C}{N}$ caused by truncation $\trunc{\mu}{N}$.
\begin{equation}
    \label{eq:trunc-obj}
    \resid{C}{N} = C - \trunc{C}{N} = C(\mu) - C(\trunc{\mu}{N})
\end{equation}
\begin{align}
    \label{eq:os-pca-problem}
    \begin{aligned}
    & \pcset{N} = \arg \min_{\pc{}} \mean{\norm{\resid{\mu}{N}}^2+\varepsilon \resid{C}{N}^2} ~\forall N \\
    &\text{s.t.}~ \neprod[\Theta]{\pc{i}}{\pc{j}} = \delta_{ij} ,~ \varepsilon \ge 0
    \end{aligned}
\end{align}
We also propose that it is sufficient to consider only local properties of an objective function such as gradient to obtain more representative subspaces $\pcset{N}$ than the original principal components.

\subsubsection{Gradient-Sensitive PCA (GS-PCA)} \label{sec:gs-pca}
Using first-order Taylor decomposition, we derive an approximation for $\resid{C}{N}$, which is a linear function of decomposition residual $\resid{\mu}{N}$. In a discrete case, $J_\eta$ is a row vector of an objective gradient at some point $\eta$.
\begin{gather} \label{eq:grad-approx}
    J_\eta = \nabla C(\eta) \\
    \resid{C}{N} \approx \eprod{J_\eta}{\resid{\mu}{N}} = J_\eta \resid{\mu}{N}
\end{gather}
We derive a more generalized form of the proposed optimality condition by introducing a symmetric gradient weighting operator G
\begin{gather}
    G(x,y) = J_\eta(x) J_\eta(y) \\
    \resid{C}{N}^2 \approx \resid{\mu}{N}^T J_\eta^T J_\eta \resid{\mu}{N} = \resid{\mu}{N}^T G \resid{\mu}{N}= \norm[G]{\resid{\mu}{N}}^2\\
    \norm[\Theta]{x}^2 = \neprod[\Theta]{x}{x}
\end{gather}
and corresponding scalar product operator $W$.
\begin{equation}
    W = I + \varepsilon G
\end{equation}
\begin{align} \label{eq:os-pca-general}
    \begin{aligned}
    & \pcset{N} = \arg \min_{\pc{}} \mean{\norm[W]{\resid{\mu}{N}}^2} ~\forall N \\
    &\text{s.t.}~ \neprod[\Theta]{\pc{i}}{\pc{j}} = \delta_{ij}
    \end{aligned}
\end{align}
The essence of our solution to this problem is to assign $W$ as a non-Euclidean scalar product operator for required principal components and use its SVD to reduce the problem to the original PCA.
\begin{gather}
    \label{eq:gs-pca-transform}
    \Theta \leftarrow W = \Phi_W \Sigma_W^2 \Phi_W^T \\
    \hat{\mu} = \Sigma_W \Phi_W^T \mu \\
    \hat{\pc{i}} = \Sigma_W \Phi_W^T \pc{i}
\end{gather}
\begin{align} \label{eq:gs-pca-problem}
    \begin{aligned}
    & \pcsethat{N} = \arg \min_{\pc{}} \mean{\norm{\resid{\hat{\mu}}{N}}^2} ~\forall N \\
    &\text{s.t.}~ \neprod[]{\hat{\pc{i}}}{\hat{\pc{j}}} = \delta_{ij}
    \end{aligned}
\end{align}
Then required principal components $\pc{i}$ are obtained from $\hat{\pc{i}}$ by a linear transform.
\begin{equation}
    \label{eq:ga-pca-solution}
    \pc{i} = \Phi_W \Sigma_W^{-1} \hat{\pc{i}}
\end{equation}

\subsubsection{Approximate GS-PCA (aGS-PCA)} \label{sec:ags-pca}
Since the GS-PCA requires two SVDs, its computational cost could be considered impractical in some cases of large high-dimensional datasets. Thus, to achieve a computational efficiency of calculating gradient-sensitive principal components, the corresponding optimal decomposition can be found approximately in the framework of the Stationary Perturbation Theory (SPT). In this framework, we treat $\varepsilon$ as a small parameter, and the whole term $\varepsilon \resid{C}{N}^2$ is considered as a small perturbation.

\begin{gather}\label{eq:spt-criterion}
    0 \leq \varepsilon \norm{J_\eta}^2 \ll 1
\end{gather}
\begin{align}
    \label{eq:ags-pca-problem}
    \begin{aligned}
    & \pcset{N} = \arg \min_{\pc{}} \mean{\norm{\resid{\mu}{N}}^2+\varepsilon \norm[G]{\resid{\mu}{N}}^2}~\forall N \\
    &\text{s.t.}~ \neprod[I + \varepsilon G]{\pc{i}}{\pc{j}} = \delta_{ij}
    \end{aligned}
\end{align}
We assume that required solution $\{\pc{i}, \sigma_i\}$ is a first-order correction of the unperturbed solution $\{\pc{i}^0, \sigma_i^0\}$ considering higher-order terms insufficient.
\begin{gather}
    \pc{k} = \pc{k}^0 + \sum_j \alpha_{kj} \pc{j}^0 ,~\alpha_{kk} = 0 \\
    \sigma_k = \sigma_k^0 + \sigma_k^1
\end{gather}
After neglecting higher-order terms of the perturbed problem and introducing gradient decomposition coefficients $b_i$, the necessary condition of constrained extremum for the perturbed problem has a form of perturbed original eigenproblem for each $\pc{k}$.
\begin{gather}
    K \pc{k} + \varepsilon b_k \sum_i b_i K \pc{i} = \sigma_k \pc{k}
    \\
    b_i = \eprod{\pc{i}^0}{J_\eta}
\end{gather}
The approximate solution is obtained by a substitute of required vectors with its first-order decomposition concerning the unperturbed solution's properties.
\begin{gather}
    \label{eq:ags-pca-solution}
    \alpha_{kn}=\varepsilon b_k b_n \frac{\sigma_n^0}{\sigma_k^0 - \sigma_n^0} \\
    \sigma_n^1 = \varepsilon (b_n)^2 \sigma_n^0
\end{gather}
Detailed derivations for the result can be found in the Appendix on page \pageref{appendix}.

\subsubsection{Gradient-Sensitive Subspace Extension (eGS-PCA)} \label{sec:egs-pca}
We also introduce a fast and computationally cheap technique of gradient-sensitive extension of subspace $\{\pc{i}^0\}_{i=1}^N$. This technique relies on a geometrical interpretation of the gradient-sensitive principal components: a smaller value of $\resid{C}{N}$ can be achieved by increasing the angle between objective gradient $J_\eta$ and residual decomposition term $\resid{\mu}{N}$.
\begin{equation}
    \resid{C}{N}=\eprod{J_\eta}{\resid{\mu}{N}}=\sum_{i>N} b_i a_i
\end{equation}
Thus, we propose that any given subspace $\{\pc{i}^0\}_{i=1}^N$ could be efficiently extended with several components $\pc{n}^0, n>N$, such that the corresponding decomposition coefficients $b_n$ are greater than the rest of the other coefficients. This idea is represented in the relative change of perturbed eigenvalues.
\begin{gather}
    {\sigma_i^1} / {\sigma_i^0} =\varepsilon (b_i)^2 \\
    \pc{n}^0 : n = \arg \max_{i>N} (b_i)^2
\end{gather}

\section{Results} \label{sec:results}
For the following numerical experiments, we generated a set of two-dimensional samples of size $21\times21$ using \texttt{rsgeng2D} function of the MySimLabs MATLAB toolbox\cite{bergstrmsurface}. $441$ train samples were gained by calling \texttt{rsgeng2D(21,3,1,1)}. A test sample was generated by \texttt{rsgeng2D(21,6,1,1)}. After that, train samples $\tau$ and test sample $\tau^*$ were rescaled to obtain required vectors $\mu$ and $\mu^*$.
\begin{gather}
    \tau_{\text{min}} = \min_{r} \tau(r) ,~ \tau_{\text{max}} = \max_{r} \tau(r) \\
    K_{\text{min}} = 1 ,~ K_{\text{max}} = 100 \\
    \mu = \ln{\frac{K_{\text{max}}}{K_{\text{min}}}} \times \frac{\tau - \tau_{\text{min}}}{\tau_{\text{max}} - \tau_{\text{min}}} + \ln{K_\text{min}}
\end{gather}
Next, we performed standard PCA and computed an initial number of components $N$ using the energy criterion with a threshold of $95\%$.
\begin{gather}
    \omega(n) = {\sum_{i=1}^n \sigma_i}/{\sum_{i} \sigma_i} \label{eq:enfrac}\\
    N = \min_{1 \leq n \leq 441} n : \omega(n) \geq 0.95
\end{gather}

Hydrocarbon flow simulation $S(\mu)$ was performed with MRST \cite{lie_2019}, and parameter $\mu$ was treated as a logarithm of permeability $K$, $[K] = 1~\text{milli-Darcy}$.
\begin{equation}
    K(\mu) = e^{\mu}
\end{equation}
The recovery setup was a five-point system of four production wells and one injection well with constant borehole pressure as a control state. Production wells were placed at the corner cells of the computational grid, and injection well was placed at the central cell.

Prior observations $S_0$ were calculated for truncated test field $\trunc{\mu}{2N}^*$ to study a contribution of principal components of not very high spatial frequencies. Thus, the whole search area was a subspace of $2N$ first standard principal components, and vector $\trunc{\mu}{2N}^*$ was treated as a ground truth. Also, the global minimum of the objective equals zero.
\begin{gather}
    \mu^* \leftarrow \trunc{\mu}{2N}^* = \sum_{i=1}^{2N} a_i^* \pc{i}^0 \\
    S_0 = S(\mu^*)
\end{gather}

In the following experiments, we set a standard approximation $\trunc{\mu}{N}^*$ as a trial point $\eta$ for gradient calculation.
\begin{equation}
    \eta = \trunc{\mu}{N}^* = \sum_{i=1}^N a_i^* \pc{i}^0
\end{equation}
An objective function gradient can be fully decomposed into $2N$ principal components.
\begin{equation}
    J_\eta^T = \sum_{i=1}^{2N} b_i \pc{i}^0
\end{equation}
We used two finite difference approximations of the gradient $J_\eta$: the central difference approximation $J_\eta^{(1)}$ similarly to \cite{kaleta2011model} and a two-point approximation in the direction to the ground truth $J_\eta^{(2)}$.

\begin{equation}\label{eq:b1}
   b_i^{(1)} = \frac{C(\eta + \Delta a_i \pc{i}^0) - C(\eta - \Delta a_i \pc{i}^0)}{2 \Delta a_i} 
\end{equation}
\begin{equation}\label{eq:J2}
   (J_\eta^{(2)})^T = - \frac{C_N^*}{\norm{\resid{\mu}{N}^*}} \times \frac{\resid{\mu}{N}^*}{\norm{\resid{\mu}{N}^*}}
\end{equation}
\begin{equation}\label{eq:b2}
b_i^{(2)} = \left\{
\begin{array}{l}
     0 ,~ 1 \leq i \leq N  \\
     -\dfrac{C_N^*}{\norm{\resid{\mu}{N}^*}} \times a_i^*,~ i > N
\end{array}
\right.
\end{equation}
The second approximation $J_\eta^{(2)}$ cannot be obtained in typically applied cases, and it only served the purpose of a thorough exploration of gradient-sensitive subspace properties.

\subsection{Experiment: Training Scores of Algorithms}\label{sec:train}
\begin{figure}[h]
\centering
\includegraphics[width = 1\linewidth]{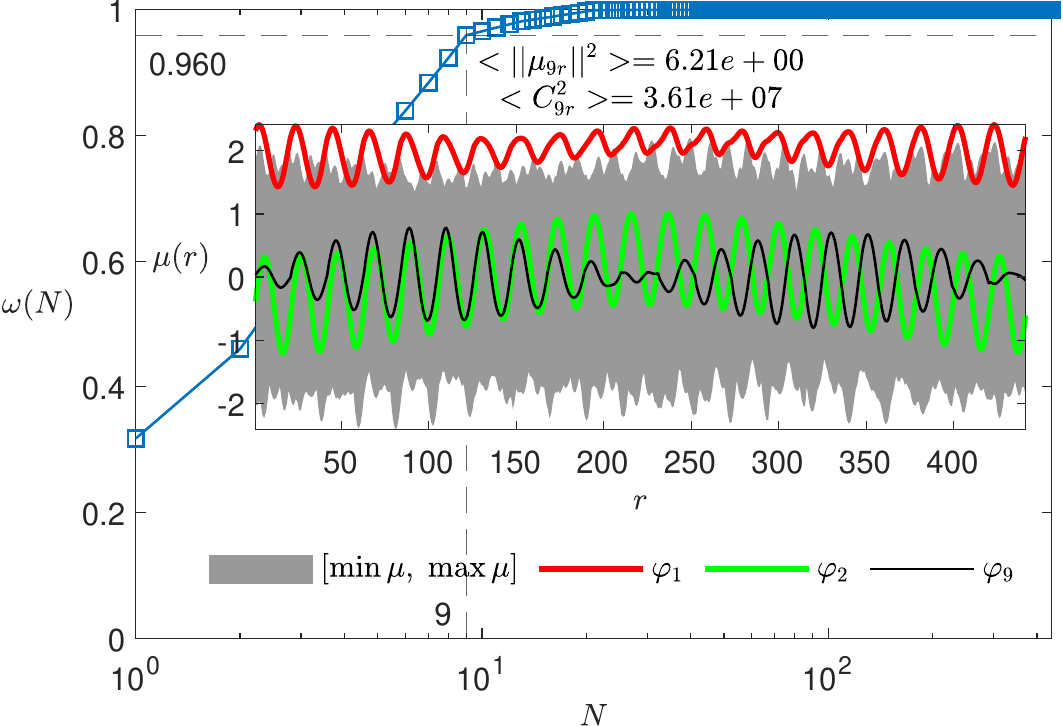}
\caption{PCA summary. This figure represents singular values $\sigma_i$~(\ref{eq:pca-svd}) through the function $\omega(N)$~(\ref{eq:enfrac}), highlighting the value of $\omega(9)$. For each feature $r$, we provide dataset minimum/maximum and vertically scaled principal components $\pc{i}$.}
\label{fig:pcaplot}
\end{figure}

\begin{figure}[h]
\centering
\includegraphics[width = 1\linewidth]{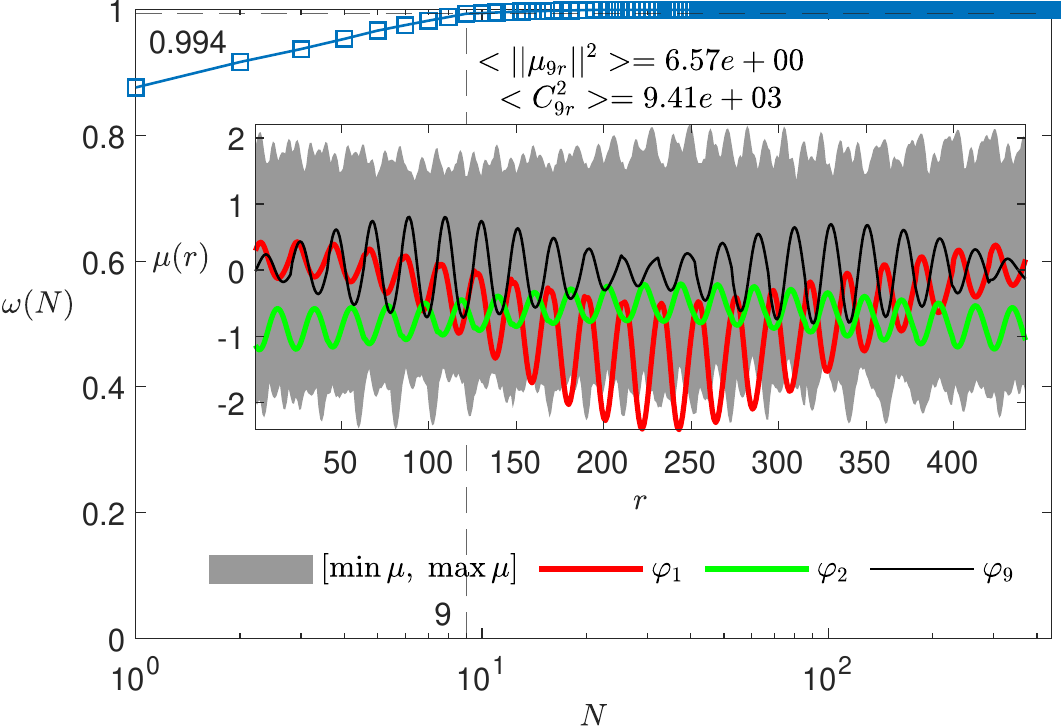}
\caption{GS-PCA summary. This figure represents singular values $\sigma_i$~(\ref{eq:pca-svd}) through the function $\omega(N)$~(\ref{eq:enfrac}), highlighting the value of $\omega(9)$. For each feature $r$, we provide dataset minimum/maximum and vertically scaled principal components $\pc{i}$.}
\label{fig:gspcaplot}
\end{figure}

The first experimental set was designed to observe how the number of components $N_1$ and the contribution of gradient-sensitivity affects subspace energy $\omega(N_1)$, loss function terms $\mean{\norm{\resid{\mu}{N_1}}^2}$ and $\mean{\resid{C}{N_1}^2}$, and shape of principal components $\pc{}$.
In this set, we computed various subspaces $\pcset{N}$ observing both $\mean{\norm{\resid{\mu}{N_1}}^2}$ and $\mean{\resid{C}{N_1}^2}$. Such observations can be interpreted as encoding scores for particular train data, objective function, and algorithm. Consequently, we treated $\resid{C}{N_1}$ as corresponding linear approximations with gradient $J_\eta^{(1)}$. The eGS-PCA algorithm was sequentially performed $(N_1-N)$ times for standard principal components to reach a required number of subspace dimensions. Results for the set are given in the Table~\ref{tab:train}. Visualization of principal components $\pc{i}$ and singular values $\sigma_i$ are given in Fig.~\ref{fig:pcaplot} and Fig.~\ref{fig:gspcaplot}.
\begin{table}[h]
\caption{Encoding scores for the train data. $J_\eta = J_\eta^{(1)}$, $\varepsilon \norm{J_\eta}^2 = 10^2$.}
\label{tab:train}
\centering
\begingroup
\setlength{\tabcolsep}{4pt}
\renewcommand{\arraystretch}{1.2}
\begin{tabular}{|c|c|r|c|c|c|}
\hline
\# & $N_1$ & Algorithm & $\omega(N_1)$ & $\mean{\norm{\resid{\mu}{N_1}}^2}$ & $\mean{\resid{C}{N_1}^2}$ \\
\hline
1 & \multirow{3}{*}{N}    &     PCA & $0.960$ & $6.21\times10^0$ & $3.61\times10^7$\\
2 &                       &  GS-PCA & $0.994$ & $6.57\times10^0$ & $9.41\times10^3$\\
3 &                       & aGS-PCA & $0.973$ & $7.68\times10^0$ & $1.96\times10^4$\\\cline{1-6}
4 & \multirow{4}{*}{1.5N} &     PCA & $0.984$ & $2.40\times10^0$ & $1.70\times10^7$\\
5 &                       &  GS-PCA & $0.998$ & $2.46\times10^0$ & $1.19\times10^2$\\
6 &                       & aGS-PCA & $0.988$ & $2.64\times10^0$ & $1.39\times10^2$\\
7 &                       & eGS-PCA & $0.977$ & $3.50\times10^0$ & $2.58\times10^6$\\\cline{1-1}\cline{3-6}
\hline
\end{tabular}
\endgroup
\end{table}


\subsection{Experiment: Projections onto Gradient-Sensitive Subspaces} \label{sec:test}
\begin{figure}[hb!]
\centering
\includegraphics[width=1\linewidth]{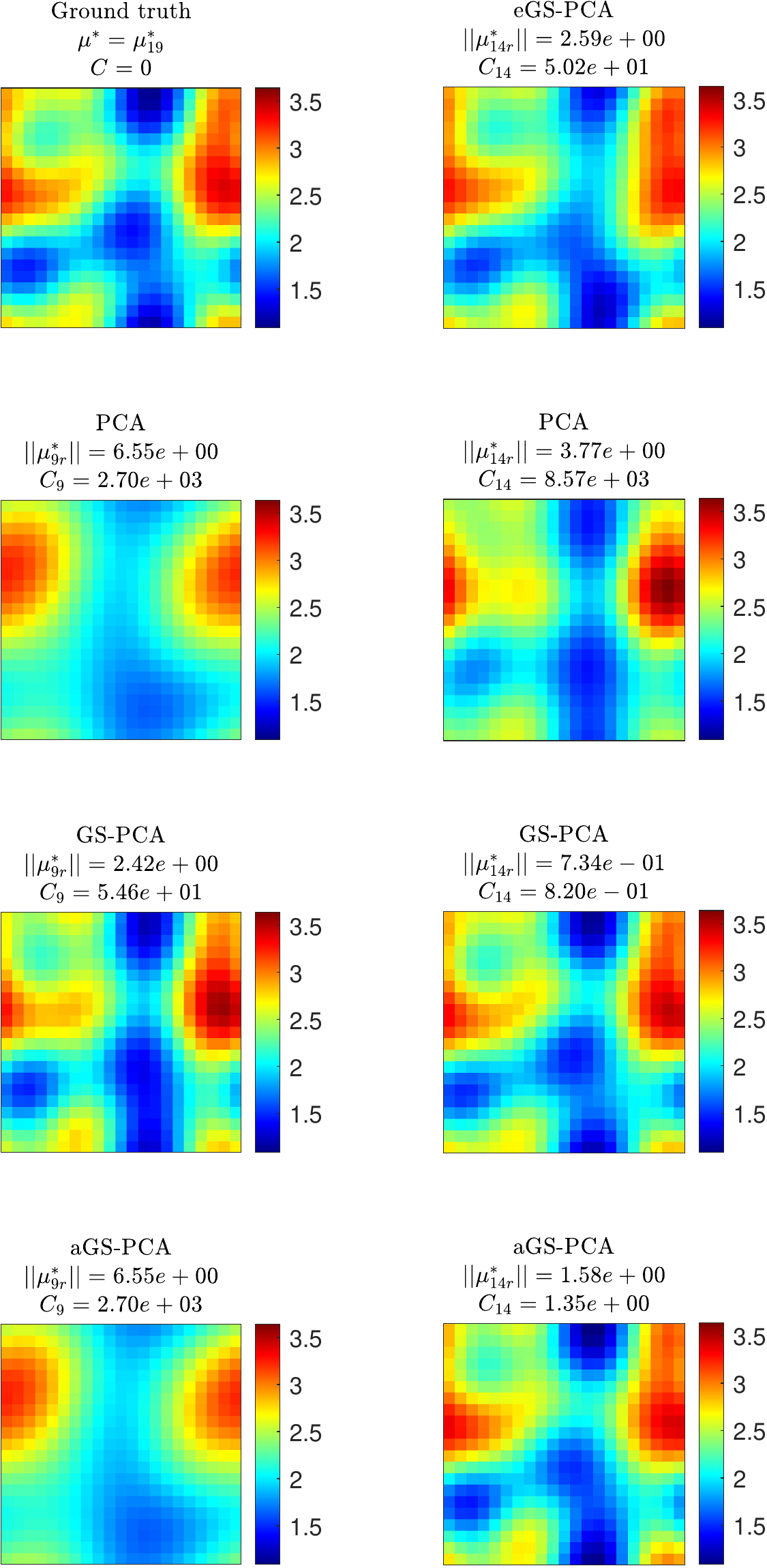}
\caption{Projections of $\mu^*$ onto principal components from Table~\ref{tab:test}. Case $J_\eta = J_\eta^{(2)}$}
\label{fig:j2} 
\end{figure}

\begin{figure}[hb!]
\centering
\includegraphics[width=1\linewidth]{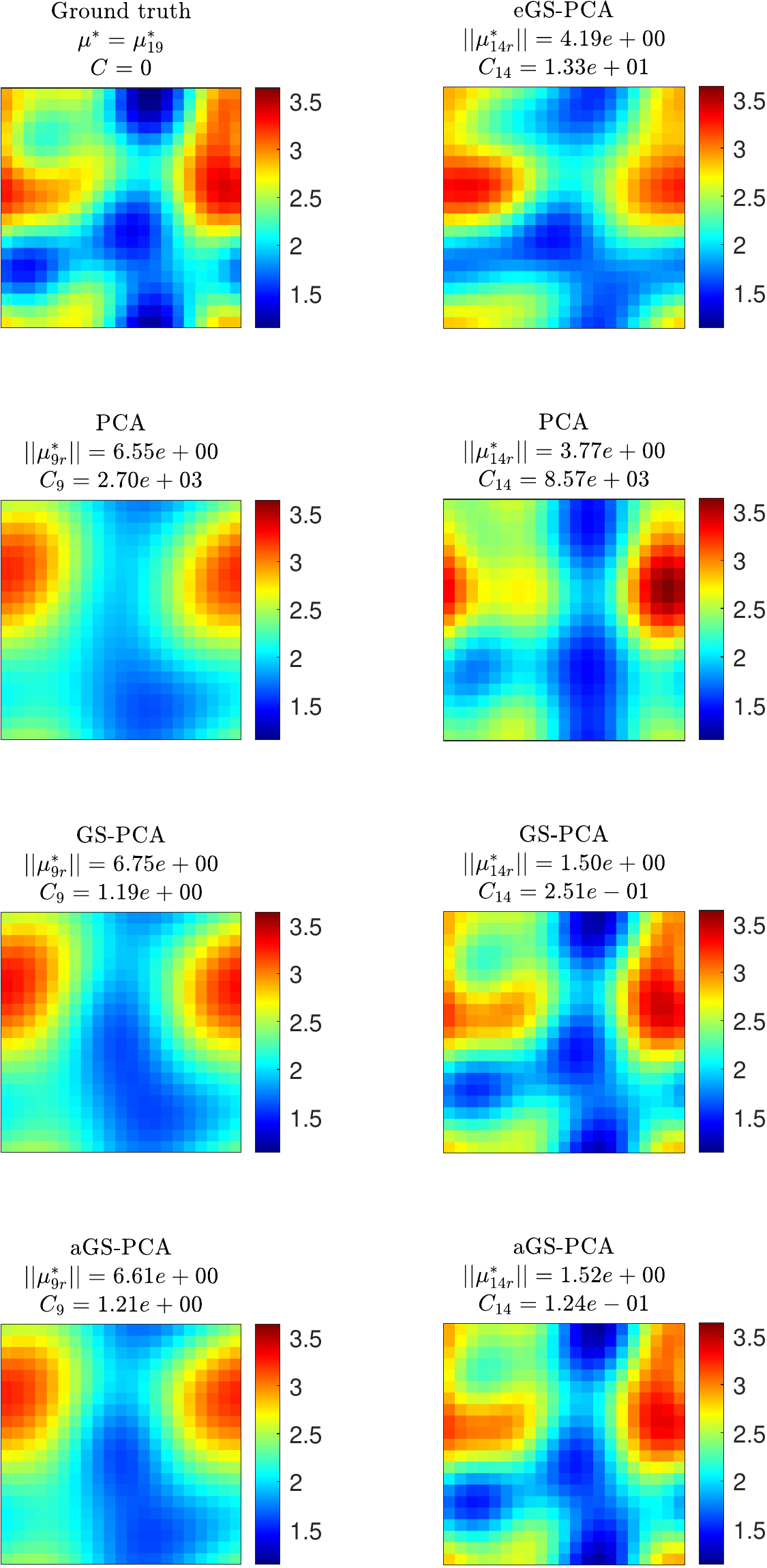}
\caption{Projections of $\mu^*$ onto principal components from Table~\ref{tab:test}.Case $J_\eta = J_\eta^{(1)}$}
\label{fig:j1}
\end{figure}

The second experimental set was designed to study the influence of GS-PCA algorithms and a gradient $J_\eta$ direction on the descriptiveness of derived subspaces regarding ground truth $\mu^*$.
In this set, we projected the test sample $\mu^*$ onto subspaces of different principal components obtaining corresponding truncation $\trunc{\mu}{N_1}^*$ and objective function value $\trunc{C}{N_1}^* = - \resid{C}{N_1}^* $. We applied both approximations of a gradient $J_\eta^{(1)}$ and $J_\eta^{(2)}$, which had different directions.
\begin{equation}
    \frac{\eprod{J_\eta^{(1)}}{J_\eta^{(2)}}}{\norm{J_\eta^{(1)}} \norm{J_\eta^{(2)}}} = 0.22
\end{equation}
Results in Table~\ref{tab:test} provide values of residuals $\norm{\resid{\mu}{N_1}^*}$ and $\norm{\resid{C}{N_1}^*}$ as truncation scores of the test sample. We also provide a visualization of the truncations in Fig.~\ref{fig:j2} and Fig.~\ref{fig:j1}.

\begin{table}[h!]
\centering
\caption{Encoding scores for the test sample. $\varepsilon \norm{J_\eta}^2 = 10^2$. Scores for both gradient approximations are provided.}
\label{tab:test}
\begingroup
\setlength{\tabcolsep}{2pt}
\renewcommand{\arraystretch}{1.4}
\begin{tabular}{|c|c|r|c c|c c|}
\hline
\multirow{1}{*}{\#} & \multirow{1}{*}{$N_1$} & \multirow{1}{*}{Algorithm} & \multicolumn{2}{c|}{$\norm{\resid{\mu}{N_1}^*}$} & \multicolumn{2}{c|}{$\norm{\resid{C}{N_1}^*}$} \\\cline{4-7}
&&&$J_\eta^{(1)}$&$J_\eta^{(2)}$&$J_\eta^{(1)}$&$J_\eta^{(2)}$\\\hline
1 & \multirow{3}{*}{N}      &   PCA & \multicolumn{2}{c|}{$6.55\times10^0$} & \multicolumn{2}{c|}{$2.70\times10^3$}\\
2 &                       &  GS-PCA & $6.75\times10^0$ & $2.42\times10^0$ & $1.19\times10^0$ & $5.46\times10^1$ \\
3 &                       & aGS-PCA & $6.61\times10^0$ & $6.55\times10^0$ & $\nexp{1.21}{0}$ & $\nexp{2.70}{3}$\\\cline{1-7}
4 & \multirow{4}{*}{1.5}    &   PCA & \multicolumn{2}{c|}{$3.77\times10^0$} & \multicolumn{2}{c|}{$8.57\times10^3$}\\
5 &                       &  GS-PCA & $1.50\times10^0$ & $7.34\times10^{-1}$ & $2.51\times10^{-1}$ &$8.20\times10^{-1}$\\
6 &                       & aGS-PCA & $1.52\times10^0$ &$\nexp{1.58}{0}$& $1.24\times10^{-1}$ &$\nexp{1.35}{0}$\\
7 &                       & eGS-PCA & $4.19\times10^0$ &$\nexp{2.59}{0}$& $1.33\times10^{1}$&$\nexp{5.02}{1}$\\\cline{1-1}\cline{3-7}
\hline
\end{tabular}
\endgroup
\end{table}

\section{Discussion} \label{sec:disc}
The results of the first set of experiments confirm that all three proposed GS-PCA algorithms are based on the correct derivations and assumptions. Since the term C represents a regularization, its smaller values given in Table~\ref{tab:test} reveal an expected effect for the proposed gradient-sensitive decompositions. That is consistent with the geometrical interpretation of the regularization: GS-PCA orients residuals orthogonally to the gradient $J_\eta$, which means that any truncation $\mu_{N_1}$ in the local area of the trial point $\eta$ less significantly affects a corresponding value of an objective function $C_{N_1}$. This circumstance also implies that a gradient-sensitive subspace could provide better local convergence of an objective due to such gradient-aligned orientation.

The regularization causes an insignificant increase of the approximation error $<||\mu_{N_1 r}||^2>$, which is proportional to the value of sensitivity parameter $\varepsilon$. Being a first-order SPT solution, aGS-PCA achieves almost the same train metrics as GS-PCA even with a sufficiently large magnitude of $\varepsilon$. eGS-PCA also demonstrates a competitive train score. We also observe a deformation of gradient-sensitive principal components as shown in Fig.~\ref{fig:pcaplot} and Fig.~\ref{fig:gspcaplot}.

Results of the second set of experiments demonstrate that the gradient-awareness of GS-PCA can uncover sufficient spatial patterns of $\mu^*$ underrepresented either in train data or in few first principal components of PCA. For the approximation $J_\eta^{(2)}$, which is aligned with a direction to the global minimum $\mu^*$, residual term $||\mu_{N_1 r}^*||$ of GS decompositions are sufficiently smaller as well as an objective function residual $||C_{N_1 r}^*||$. A visualization in Fig.~\ref{fig:j2} reveals that gradient-sensitive projections have more acceptable spatial properties compared to that of PCA projections of the same dimensionality.

If a gradient is either orthogonal or aligned to the original PCA subspace, the first $N$ components of aGS-PCA remain unperturbed according to the expression for the transform matrix $\alpha_{kn}$. In such a case, a dimensionality increase is necessary.

As we partially noted in the previous section, a direction given by $J_\eta^{(2)}$ cannot be reliably obtained for sufficiently complex non-convex objective functions, which are typical for practical cases. If having a significantly different direction, the 'proper' gradient approximation $J_\eta^{(1)}$ may not be able to supply notable improvements in comparison with the standard PCA, as shown in Table~\ref{tab:test} and Fig.~\ref{fig:j1}.

Although, we provide an analysis of the mentioned limitation and some other limitations and advantages associated with the developed algorithms. We also propose some possible overcomes and possible directions for further research.

\paragraph{A mismatch between local and global properties.}
We suppose that iterative recalculation of GS-PCA during inverse problem solving may be an overcome for the mentioned mismatch between $J_\eta^{(1)}$ and $J_\eta^{(2)}$ and also a way to explore a wider area than using PCA parametrization of the same dimensionality. Furthermore, considering multiple points $\eta$ may be a way to account for the nonlinearity of an objective, although we yet observed an acceptable improvement of subspace properties with a single trial point.

\paragraph{Algorithm extensions.}
The developed algorithms can be naturally extended to multi-objective problems by adding multiple corresponding regularization terms. Since the regularization affects a loss function $<||\mu_{N_1 r}||^2 + \varepsilon C_{N_1 r}^2>$ to be minimized, a consideration of objective function properties can be translated to any other advanced data-driven encoding technique such as kernel-PCA or autoencoders\cite{LALOY2017387}, but possible computational efficiency is debatable. 

\paragraph{Computational costs.}
At the same time, the GS-PCA algorithm requires one additional SVD to that of PCA, and both aGS-PCA and eGS-PCA algorithms imply much less arithmetical operations without a need to re-execute an SVD at all.

\paragraph{Gradient optimization issues.} 
Accuracy of the gradient approximation $J_\eta^{(1)}$ depends on its formula and discrete step $\Delta a$. It is also a computationally expensive way to obtain a gradient calculated along with enough number of principal components. A possible solution is to exploit the adjoint gradient if required access to the model $S(\mu)$ is provided. Nevertheless, the area of applicability of OS-PCA is not limited by gradient optimization. A possible direction of further research is to study an efficient gradient evaluation and apply OS-PCA algorithms to other approaches, such as stochastic optimization and ensemble methods.

\section{Conclusion} \label{sec:concl}
A novel algorithm for objective-sensitive principal component analysis of random fields was developed. This particular PCA-based set of methods called gradient-sensitive PCA (GS-PCA) is an extension of standard PCA with information about an objective function gradient involved in the parameterization process. The gradient-sensitive parameterization can be done using one of three suggested algorithms of GS-PCA, which provide sufficient improvement in extremum exploration. Although the approach has been tested on synthetic samples, applicability on more practical models is expected, since real data have structures underrepresented in first principal components. Given an objective gradient, methods of GS-PCA provide practically low computational costs, since GS-PCA requires two SVDs, aGS-PCA is calculated by only a few linear transforms of initial basis, and eGS-PCA implies only one comparison for each additional dimension.
Although in this paper, the set of algorithms is initially developed for gradient optimization, its implementation simplicity and relatively low computational cost allow to effortlessly improve the overall quality of inverse problem solution using black-box optimization and ensemble methods. This statement is also motivated by local convergence of gradient algorithms, which significantly limit the performance of history matching since the process is easily trapped into local minima. Approaches for efficient usage within non-gradient optimization are among the possible directions for further research.
Despite that GS-PCA was considered in the case of history matching problem, it is applicable for any other inverse problems and can be extended for usage within any other encoding technique.


%
%

\bibliographystyle{spmpsci}      
\bibliography{biblio}   


\newpage
\section{Appendix: aGS-PCA evaluation} \label{appendix}
\subsection{Perturbed eigenproblem evaluation} \label{perturbed:problem}
To obtain an approximate solution for GS-PCA, we consider the objective-sensitive optimal decomposition problem with a sensitivity parameter $\varepsilon$, which satisfies a small perturbation criterion.
\begin{align}
    \begin{aligned}
    & \pcset{N} = \arg \min_{\pc{}} \mean{\norm[W]{\resid{\mu}{N}}^2} \\
    &\text{s.t.}~ \neprod[W]{\pc{i}}{\pc{j}} = \delta_{ij} ,~ \varepsilon \ge 0
    \end{aligned}
\end{align}
\begin{gather}
    W = I + \varepsilon J^T J \\
    0 \leq \varepsilon \norm{J_\eta}^2 \ll 1
\end{gather}
Next, we express a residual norm $\norm[W]{\resid{\mu}{N}}^2$ in terms of principal components $\pc{i}$ and decomposition coefficients $a_i$.
\begin{equation}
    \norm[W]{\resid{\mu}{N}}^2 = \sum_{i>N} \sum_{j>N} a_i a_j \neprod[W]{\pc{i}}{\pc{j}}
\end{equation}
Assuming that the required solution is the first-order correction of standard principal components, we neglect higher-order terms of the scalar product $\neprod[W]{\pc{i}}{\pc{j}}$.
\begin{gather}
    \pc{k} = \pc{k}^0+\pc{k}^1\\
    \eprod{\pc{i}^0}{J_\eta} = b_i \\
    \neprod[W]{\pc{i}}{\pc{j}} \approx \eprod{\pc{i}}{\pc{j}} + \varepsilon b_i b_j
\end{gather}
After that, we express the mean product $\mean{a_i a_j}$ in terms of principal components $\pc{i}$, covariance $K$, scalar product operator $W$, and derive a corresponding expression for the mean residual norm $\mean{\norm[W]{\resid{\mu}{N}}^2}$.
\begin{gather}
    \mean{a_i a_j} = \pc{i}^T W \mean{\mu \mu^T} W \pc{j} = \neprod[WKW]{\pc{i}}{\pc{j}} \\
    \mean{\norm[W]{\resid{\mu}{N}}^2} \approx \sum_{i>N} \sum_{j>N} (\eprod{\pc{i}}{\pc{j}} + \varepsilon b_i b_j) \neprod[WKW]{\pc{i}}{\pc{j}}
\end{gather}
To achieve a sufficient simplicity of the approximate decomposition problem, we additionally neglect first-order perturbations in the obtained scalar product matrix $WKW$ and the problem constraint.
\begin{gather}
    WKW = K + \varepsilon (GK + KG) + \varepsilon^2 GKG \approx K \\
    \neprod[I+\varepsilon G]{\pc{i}}{\pc{j}} = \eprod{\pc{i}}{\pc{j}} + \varepsilon \neprod[G]{\pc{i}}{\pc{j}} \approx \eprod{\pc{i}}{\pc{j}} = \delta_{ij}
\end{gather}
\begin{align}
    \begin{aligned}
    & \pcset{N} = \arg \min_{\pc{}} \sum_{i>N} \sum_{j>N} (\eprod{\pc{i}}{\pc{j}} + \varepsilon b_i b_j) \neprod[K]{\pc{i}}{\pc{j}} \\
    &\text{s.t.}~ \eprod{\pc{i}ц}{\pc{j}} = \delta_{ij}
    \end{aligned}
\end{align}
Such constraint optimization implies the minimization of a corresponding Lagrangian. 
\begin{align}
    \begin{aligned}
    &F_i = \sum_{j>N}(\delta_{ij} + \varepsilon b_i b_j)\neprod[K]{\pc{i}}{\pc{j}} \\
    &L = \sum_{i>N}\left[
    F_i
    - \sigma_i(\eprod{\pc{i}}{\pc{j}} - 1) 
    \right] \rightarrow \min
    \end{aligned}
\end{align}
Evaluating the necessary condition of constraint extremum, 
\begin{gather}
\frac{\partial}{\partial \pc{k}} \neprod[\Theta]{\pc{i}}{\pc{j}} = 2 \delta_{jk} \Theta \pc{i} \\
\frac{\partial L}{\partial \pc{k}} = 2 (K\pc{k} + \varepsilon b_k \sum_i b_i K \pc{i} - \sigma_k \pc{k}) = 0
\end{gather}
we derive a perturbed eigenproblem associated with the aGS-PCA.
\begin{equation}
    K\pc{k} + \varepsilon b_k \sum_i b_i K \pc{i} = \sigma_k \pc{k}
\end{equation}

\subsection{Perturbed eigenproblem solution} \label{perturbed:solution}
A first-order term $\pc{i}^1$ can be represented as a linear combination of unperturbed principal components given by a transform matrix $\alpha_{ij}$. 
\begin{gather}
    \pc{k}^1 = \sum_j \alpha_{kj} \pc{j}^0 ,~ \alpha_{kk} = 0 \\
    \sigma_k = \sigma_k^0 + \sigma_k^1
\end{gather}
Given the known properties of unperturbed eigenvectors, we substitute decomposed vectors $\pc{}$ in the perturbed eigenproblem and project the expression onto an unperturbed vector $\pc{n}^0$.
\begin{gather}
    K\pc{k}^0 = \sigma_k^0 \pc{k}^0 \\
    \sum_{j} \alpha_{kj} \sigma_j^0 \pc{j}^0
    + \varepsilon b_k \sum_{i} b_i \sigma_i^0 \pc{i}^0
    = \sigma_k^0  \sum_{j} \alpha_{kj} \pc{j}^0 + \sigma_k^1 \pc{k}^0 \\
    \alpha_{kn} \sigma_n^0
    + \varepsilon b_k b_n \sigma_n^0
    = \sigma_k^0  \alpha_{kn} + \sigma_k^1 \delta_{kn}
\end{gather}
Finally, the required expressions for the unknowns $\sigma_n^1$ and $\alpha_{kn}$ are obtained from two cases of a relation between indexes $k$ and $n$.
\begin{gather}
    k = n : \varepsilon b_n b_n \sigma_n^0 = \sigma_n^1 \\
    k \neq n : \alpha_{kn} \sigma_n^0 + \varepsilon b_k b_n \sigma_n^0 = \sigma_k^0  \alpha_{kn}
\end{gather}

\end{document}